# DiffGrad for Physics-Informed Neural Networks

**Jamshaid Ul Rahman[1,2,*], Nimra[2]**

[1]School of Mathematical Sciences, Jiangsu University 301 Xuefu road, Zhenjiang 212013, China.
[2]Abdus Salam School of Mathematical Sciences, Govt College University, Lahore 54600, Pakistan
*Corresponding author: Jamshaid@sms.edu.pk

**Abstract:**

Physics-Informed Neural Networks (PINNs) are regarded as state-of-the-art tools for addressing highly nonlinear problems based on partial differential equations. Despite their broad range of applications, PINNs encounter several performance challenges, including issues related to efficiency, minimization of computational cost, and enhancement of accuracy. Burgers' equation, a fundamental equation in fluid dynamics that is extensively used in PINNs, provides flexible results with the Adam optimizer that does not account for past gradients. This paper introduces a novel strategy for solving Burgers' equation by incorporating DiffGrad with PINNs, a method that leverages the difference between current and immediately preceding gradients to enhance performance. A comprehensive computational analysis is conducted using optimizers such as Adam, Adamax, RMSprop, and DiffGrad to evaluate and compare their effectiveness. Our approach includes visualizing the solutions over space at various time intervals to demonstrate the accuracy of the network. The results show that DiffGrad not only improves the accuracy of the solution but also reduces training time compared to the other optimizers.

## 1. INTRODUCTION

Deep neural network's invention has revolutionized numerous fields including computer vision [1], game theory, and natural language processing [2]. Through deep learning (DL), the landscape of tasks involving categorization, pattern recognition, and regression has been fundamentally reshaped across a wide array of application domains [3]. Deep neural networks (DNNs) can be effectively used in oscillator simulations. Oscillators, which produce periodic signals, are common in many fields such as electronics, physics, and engineering [4]. Moreover,

Deep neural networks (DNNs) can be effectively used to simulate the Selkov glycolysis model by learning the non-linear dynamics from synthetic data generated via numerical solutions of the model's differential equations [5]. However, their utilization remains relatively unexplored in the realm of scientific computing. The incorporation of machine learning techniques into scientific computing [6], particularly for dynamic problems, is a relatively recent advancement [7].

A wide range of dynamic problems can be formulated as differential equations (DEs). Various numerical and analytical approaches such as finite element, finite difference, and finite volume methods have been proposed by researchers to solve DEs [8] While these traditional methods have demonstrated considerable success, they often fall short when dealing with real-world nonlinear problems. Moreover, each traditional method has its own set of strengths, weaknesses, and areas of applicability [9]. Additionally, these methods are problem-specific and rely on structured meshes, limiting their flexibility. Certain Partial Differential Equations(PDEs) are infamously challenging to solve using conventional numerical techniques because of factors like shocks, convective dominance, or large nonlinearities, among others.

In the discipline of mathematical physics, the numerical solution of Nonlinear PDEs has always been a hot topic in recent decades. Researchers have been able to take on increasingly challenging and high-dimensional problems thanks to developments in processing capacity and methods [10,11].

Physics-Informed Neural Networks (PINNs) represent a state-of-the-art approach in scientific machine learning, specifically engineered to address complex PDEs problems [12-14]. They integrate physical laws into the learning process, enabling the model to achieve accurate and efficient solutions to these challenging equations [15-18]. According to a two-part article, in 2017, PINNs were first introduced as a new category of data-driven solutions [19, 20] which was subsequently merged and published in 2019 [21]. In this work, Raissi et al. [22, 23] present and demonstrate PINN framework for solving nonlinear Partial Differential Equations, such as the Schrödinger, Burgers, and Allen-Cahn equations. They introduce PINNs, which are capable of addressing both inverse problems (where model parameters are inferred from observational data) and forward problems (which involve predicting solutions based on governing mathematical models). By integrating physical laws directly into the neural network framework, PINNs enhance the model's generalization capabilities, offering robust and accurate solutions for various applications in science and engineering. PINNs base their modeling strategy on a combination of underlying physical laws and empirical data.

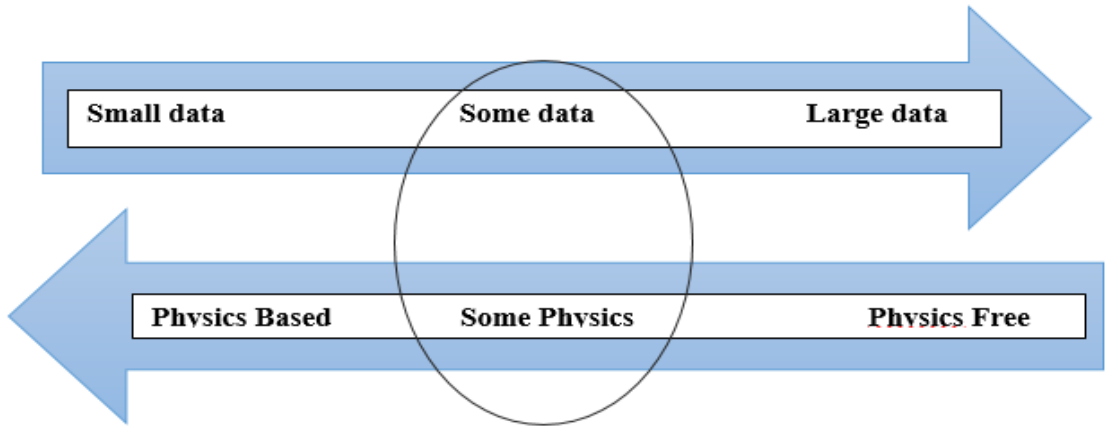

**Fig. 1**: *Schematic representation of three possible cases a dynamical problem can be categorized.*

These networks approximate PDEs solutions through training a neural network to minimize a loss function that includes terms reflecting the initial and boundary conditions across the space-time domain boundaries as well as the Partial Differential Equation (PDE) residual at particular points within the domain (called collocation points) [24]. PINNs are distinguished by their deep-learning architecture, which, upon receiving an input point within the integration domain, yields an estimated solution to a differential equation following training [25]. One of the key innovations of PINNs is their integration of a residual network that encodes the governing physics equations, allowing the model to inherently respect these physical laws while solving both forward and inverse problems efficiently [26,27,28]. The fundamental principle behind PINNs training is akin to Self-supervised approach, obviating the need for labeled data derived from previous simulations or experiments [29]. Essentially, the PINNs algorithm operates as a mesh-free technique, transforming the task of directly solving governing equations into an optimization problem centered around minimizing a loss function [30,31]. By incorporating the mathematical model into network and strengthening the loss function with a residual term taken from the governing equation, this methodology efficiently reduces the range of feasible solutions [32].

This illustrates the components of PINNs. These networks comprise residual terms from differential equations, along with initial conditions and boundary conditions. The inputs of the network, represented by variables, are converted into outputs denoted as field u, governed by parameters $\theta$ [33,34]. Then physics-informed neural network operates on the output field u, computing derivatives based on the specified equations. It also evaluates boundary/initial conditions (if not hardcoded in the neural network) and calculates observations from labeled data if available. The feedback mechanism optimizes the network parameters by minimizing the loss through the use of an optimizer with a predetermined learning rate [35,36].

PINNs consist of three primary components: a neural network (NN), a physics-informed network, and a feedback mechanism. The first component, the neural network, generates the field value u from input vector variables z. The second component acts as the functional part of PINNs by using derivatives to evaluate terms in the equation, initial conditions, and boundary conditions [37,38,39]. These two components are integrated during training using algorithmic differentiation, enabling the incorporation of physical equations into the NN. The feedback mechanism then adjusts the NN parameters $\theta$ by minimizing the loss based on a specified learning rate [40]. Depending on the circumstances, we shall make it clear throughout the conversation whether we are speaking of the functional network or the neural network that is in charge of obtaining physical information [41,42]. PINNs have been the subject of numerous review papers that have been published. These studies explore the capabilities, limitations, and practical applications of PINNs in addressing both forward and inverse problems, encompassing scenarios such as three-dimensional flows and comparative analyses with alternative machine learning methodologies [43]. In their work, Kollmannsberger et al. provide a foundational course on PINNs, delving into the fundamentals of neural networks and machine learning principles [44]. These resources contribute to a comprehensive understanding of how PINNs integrate physical constraints into neural networks, advancing their utility across various scientific and engineering disciplines. Additionally, PINNs have been contrasted with alternative approaches to PDEs solving, such as those that rely on the Feynman-Kac theorem [45]. Furthermore, integrodifferential equations (IDEs) and stochastic differential equations (SDEs) can be solved using PINNs [46]. Because PINNs can learn PDEs, they have a number of advantages over traditional techniques. These mesh-free techniques enable on-demand computation of solutions after training and utilize analytical gradients to ensure differentiability of solutions, facilitating their integration into optimization and uncertainty quantification frameworks. PINNs offer a simple method for applying the same optimization framework to address both forward and inverse problems [47,48]. PINNs demonstrate versatility in handling inverse problems, including the characterization of fluid flows from sensor data with minimal code adjustments, alongside their proficiency in solving forward differential equations enhancing their capability to tackle complex inverse problems. In inverse design, PINNs can enforce PDEs as stringent constraints, making them highly effective for tasks such as high-performance PINNs, where accurate modeling of complex systems is crucial [49]. Their ability to operate in domains featuring intricate geometries or high-dimensional spaces makes them particularly adept for challenging tasks like constrained optimization and sophisticated numerical simulations. PINNs thus offer a robust framework for integrating physical laws into machine learning models, paving the way for innovative applications across diverse scientific and engineering domains. [50].

In this work, we introduce a novel strategy for solving Burgers' equation by incorporating DiffGrad with PINNs, a technique that leverages the difference between current and immediately applying prior gradient information to boost the performance. A comprehensive analysis is conducted using promising optimizers including Adam, Adamax, RMSprop, and DiffGrad to evaluate and compare their effectiveness.

2. **METHODOLOGY**

The methodology involves creating an approximation using a neural network $u_\theta(t, x)$ to represent the solution $u(t, x)$ of a nonlinear PDE within a specific domain $D$ and time range $[0, T]$. In this context, $u_\theta : [0, T] \times D \to R$ represents a function represented by a neural network with parameters $\theta$. The approach involves a continuous-time formulation for the parabolic PDE, as discussed in (Raissi et al., 2017 (Part I)), where the focus is on the residual of the neural network approximation $u_\theta(t, x)$ and $r_\theta(t, x)$.

This residual is defined as

$$r_\theta(t, x) := \partial_t u_\theta(t, x) + N[u_\theta](t, x) \qquad (1)$$

where $N[u_\theta](t,x)$, represents the nonlinear component of the PDE.

To incorporate this PDE residual $r_\theta$ into a loss function for minimization, Physics-Informed Neural Network(PINN) necessitate additional differentiation techniques to accurately compute differential operators $\partial_t u_\theta$ and $N[u_\theta]$. Consequently, the PINN term shares the same parameters as the original network $u_\theta(t, x)$ while respecting underlying physics of the nonlinear PDE. The parameters shared between the neural networks $h(t, x)$ and $f(t, x)$ are optimized by minimizing the mean squared error loss. Both type of derivatives can be efficiently computed through automatic differentiation using modern machine learning libraries such as TensorFlow or PyTorch.

The beginning and boundary value problems are now solved using the PINNs technique by minimizing the loss function.

$$\emptyset_\theta(X) := \emptyset_\theta^r(X_r) + \emptyset_\theta^0(X_0) + \emptyset_\theta^b(X_b) \qquad (2)$$

here X stands for both the loss function and the gathering of training data. The following terms are contained in $\emptyset_\theta$: The mean squared residual,

$$\emptyset_\theta^r X_r := \frac{1}{N_r}\sum_{j=1}^{N_r} |r_\theta(t_j^r, x_j^r)|^2 \qquad (3)$$

At several collocation points $X_r = \{(t_j^r, x_j^r)\}_{j=1}^{N_r} \subset (0, T] \times D$, where $r_\theta$ is our physics-informed network and $N_r$ is the total number of collocation points.

The mean squared error related to the initial and boundary conditions is determined by averaging the squared differences between the predicted values from the neural network approximation $u_\theta$ and the actual initial conditions at points $X_0$, and the boundary conditions at points $X_b$. This misfit quantifies the discrepancies in the predictions made by the neural network and the specified initial and boundary conditions

$$\emptyset_\theta^0(X_0) = \frac{1}{N_0}\sum_{j=1}^{N_0} |u_\theta(t_j^0, x_j^0) - u_0(x_i^0)|^2 \qquad (4)$$

$$\emptyset_\theta^b(X_b) = \frac{1}{N_b}\sum_{j=1}^{N_b} |u_\theta(t_j^b, x_j^b) - u_b(x_i^b)|^2 \qquad (5)$$

At several points

$$X_0 := \{(t_j^0, x_j^0)\}_{j=1}^{N_0} \subset \{0\} \times D$$

and

$$X_b := \{(t_j^b, x_j^b)\}_{j=1}^{N_b} \subset (0, T] \times \partial D$$

Where $u_\theta$ is the neural network's approximation of the solution $u: [0, T] \times D \to R$.

It's worth noting that the training dataset $X$ exclusively comprises coordinates representing both time and space.

The entire process can be visually represented through a figure to provide a clear illustration of the steps involved as shown in Fig.02.

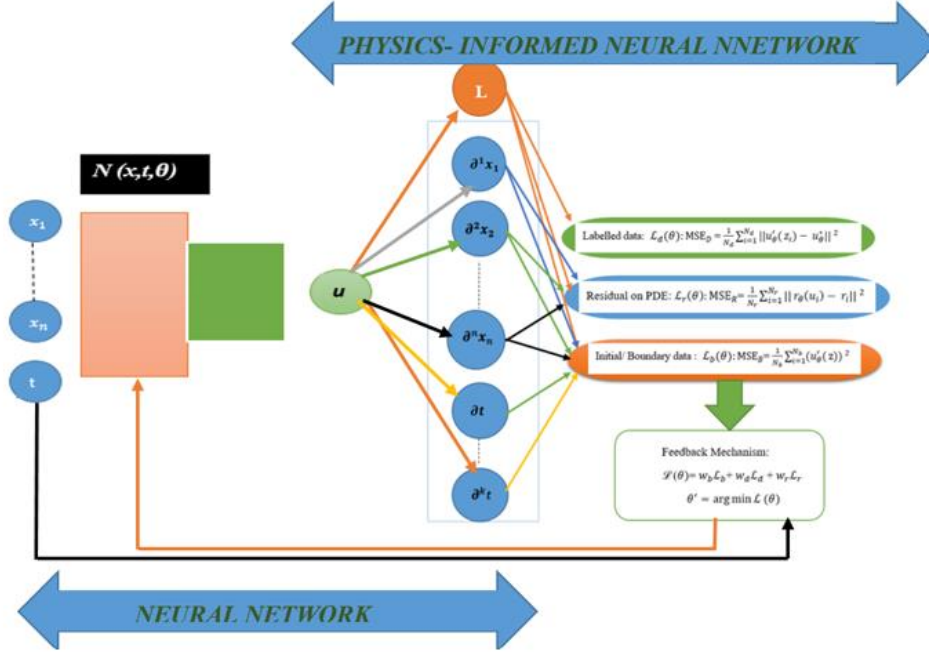

*Fig.02: Methodology employed for a physics-informed neural network (PINN) model*

3. **PROBLEM SETUP**

To exemplify the Physics-Informed Neural Networks (PINNs) methodology, we focus on one-dimensional Burgers equation within the spatial domain $D = [-1,1]$

$$\partial_t u + u\partial_x u - \left(\frac{0.01}{\pi}\right)\partial_{xx} u = 0, \quad (t,x) \in (0,1] \times (-1,1) \qquad (6)$$

$$u(0,x) = -\sin(\pi x), \quad x \in [-1,1]$$

$$u(t,-1) = u(t,1) = 0, \quad t \in (0,1]$$

The provided equation (6) describes a one-dimensional Burgers' equation incorporating Dirichlet boundary conditions and a source term. It describes the evolution of a velocity field $u(t,x)$ over time $t$ and space $w$. The equation includes convection, diffusion, and a nonlinear advection term $u\partial_x u$. The boundary conditions set the velocity to zero at both ends of the domain $x = -1$ and $x = 1$ while the initial condition $u(0,x) = -\sin(\pi x)$ defines the initial velocity distribution at $t = 0$. The source term $-\left(\frac{0.01}{\pi}\right)\partial_{xx} u$ introduces additional effects on the velocity field. This equation governs the behavior of the velocity field within the specified domain $(0,1] \times (-1,1)$ over the time interval $t \in (0,1]$.

The mentioned PDE appears in diverse fields including traffic dynamics, fluid mechanics, and gas dynamics. It can be obtained as a simplification of the Navier-Stokes equations.

## 4. ALGORITHMIC FRAMEWORK AND ARCHITECTURE

The neural network architecture cosists of the following steps.

### i. Initialization of Packages and Problem-Specific Parameters

The code utilizes TensorFlow version 2.3.0, along with the NumPy library for scientific computing. TensorFlow is used for machine learning tasks. All the optimizers utilized in the code were built-in optimizers provided by TensorFlow, except for DiffGrad. DiffGrad is a custom optimizer implemented within the code, tailored to specific requirements or experimental purposes.

### ii. Generate a set of collocation points

The initial time and boundary data points, $X_0$ and $X_b$, along with the collocation points $X_r$, are generated by randomly sampling from a uniform distribution. In our experiments, uniformly distributed data suffice for testing purpose; however, previous work (Raissi et al., 2017 (Part I)) has shown that a space-filling Latin hypercube sampling approach (Stein, 1987) can marginally improve the rate of convergence. This observation is verified by our numerical investigations. However, the code example provided in this paper employ uniform sampling consistently for the sake of simplicity. We choose to use $N_0 = N_b = 50$ and $N_f = 10000$ as our training data sizes.

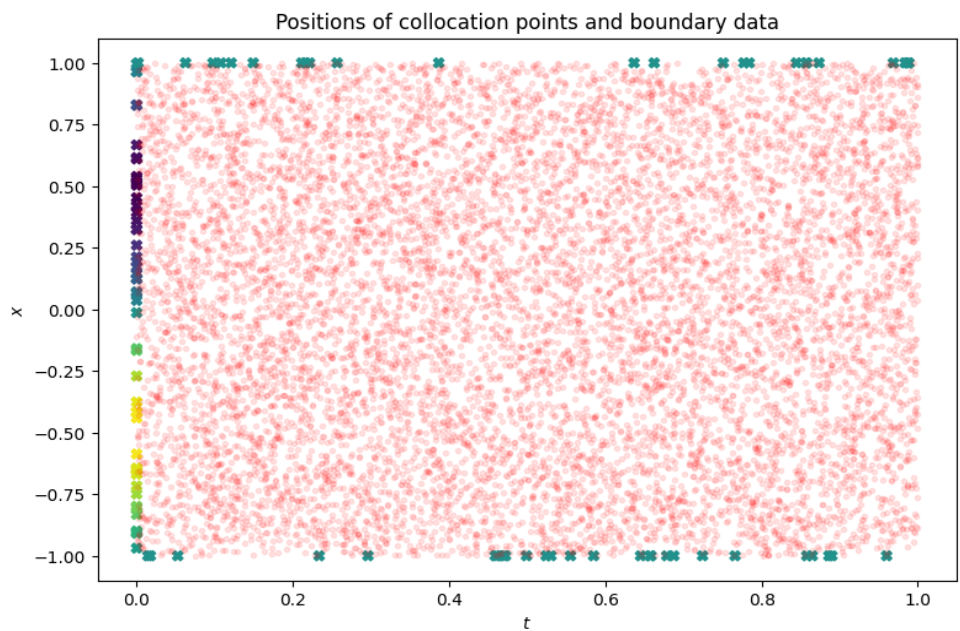

*Fig. 03: Position of collocation points and boundary data*

The Fig.03 illustrates the distribution of collocation points and boundary data utilized in the study, providing an overview of the spatial arrangement used for training the PINNs model.

### *iii.* **Establishing Neural Network Configuration**

In this example, inspired by Raissi et al. (2017, Part I), we employ a Feedforward Neural Network structured as follows:

Eight completely connected layers follow after the input is scaled element-wise to fall inside the range of $[-1,1]$ in the network architecture. Twenty neurons make up each layer, which is driven by a hyperbolic tangent activation function. There is another one output layer that is entirely connected. This setup results in a network with a total of 3021 trainable parameters distributed across its layers:

- ➢ The first hidden layer consists of 60 parameters $(2 \times 20 + 20)$.
- ➢ Each of the seven intermediate layers encompasses 420 parameters $(20 \times 20 + 20)$.
- ➢ The output layer comprises 21 parameters $(20 \times 1 + 1)$.

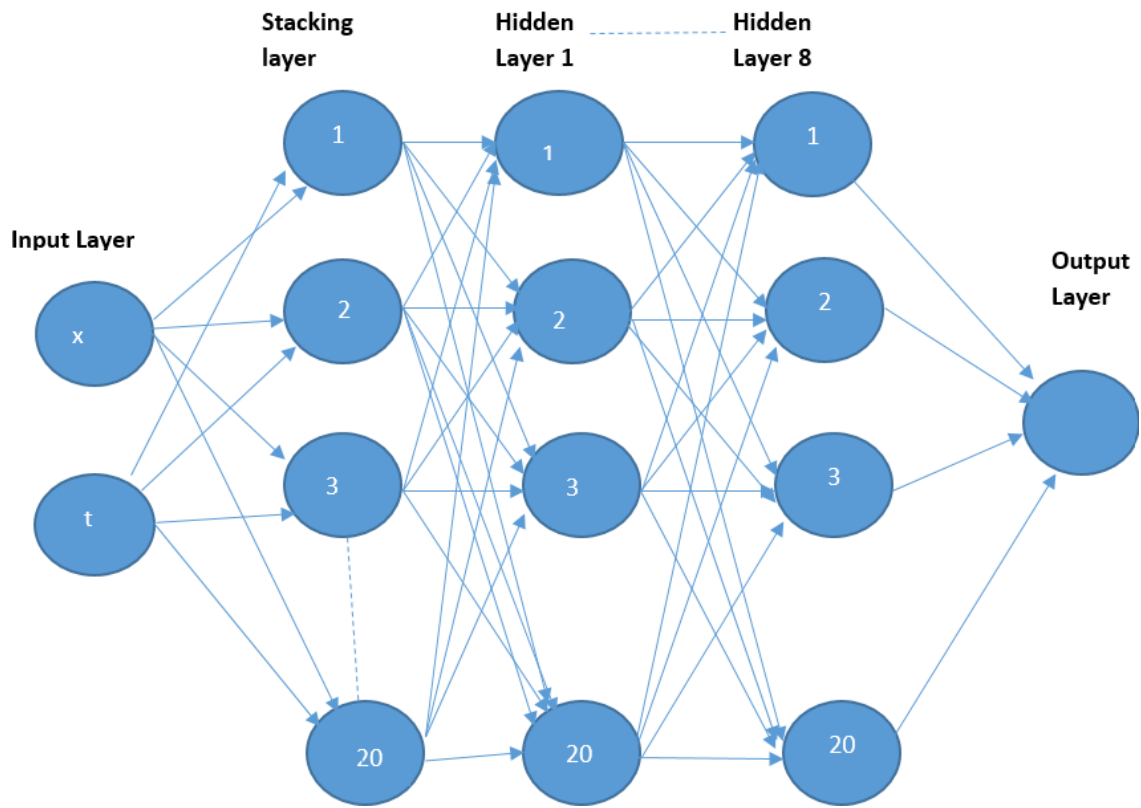

*Fig.04: Neural Network Architecture*

The Fig. 04 illustrates the schematic representation of the neural network architecture employed in our study which consists of one input layer, one stacking layer, eight hidden layers and one output layer. The connections between neurons indicate the flow of information through the network, with weights and biases associated with each connection. This architecture facilitates the mapping of input data to output predictions, enabling the network to learn and approximate solutions to Burgers' equation.

### iv. Establish Functions for Calculating Loss and Gradients:

Next we define a function which evaluates the residual using equation (1) and loss using equation (2) and initial and boundary conditions as given in equation (3) and (4) respectively.

## 5. RESULTS AND DISCUSSION

The objective of this study is to analyze the behavior of various optimizers in conjunction with PINNs for solving Burgers' equation. Specifically, we aim to investigate how different optimizers affect the convergence and accuracy of the solution obtained by PINNs.

We investigate the performance of various optimization algorithms in minimizing the loss function described above. The optimizers considered in our study include

a) DiffGrad
b) Adam
c) RMSprop
d) Adamax

Each optimizer aims to update the network weights to minimize the loss function iteratively. We employ the TensorFlow library to implement these optimization algorithms efficiently.

To train the PINNs model, we generate collocation points within the domain. These points consist of initial points $(x,t)$, boundary points, and randomly sampled collocation points within the interior of the domain as shown in Fig. 03. Starting with differential gradients (diff grad) as our optimization method, we aim to assess its effectiveness in enhancing the performance of Physics-Informed Neural Networks (PINNs) models. Next we will see the results with Adam.

The plot of loss function over epochs is under:

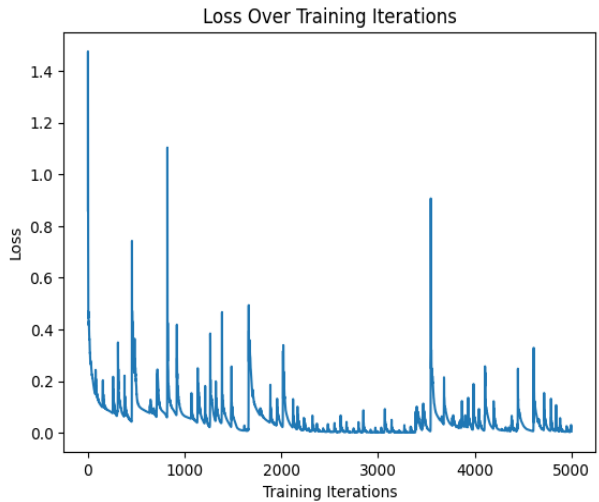 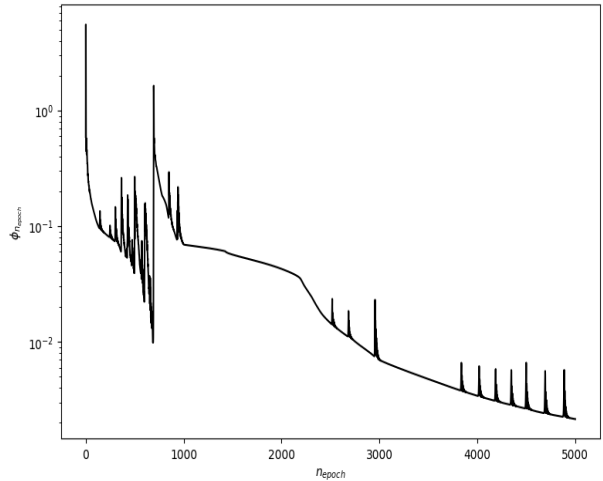

*Fig.05: Loss over training Ietrations with Optimizer DiffGrad*

*Fig.06: Loss over training Ietrations with Optimizer Adam*

The plot in Fig. 05 illustrates the evolution of the loss function over training epochs during the optimization process using the differential gradients (diff grad) algorithm. Each point on the curve represents the average loss calculated over a specific number of epochs, providing a visual representation of the convergence behavior of PINNs model. Initially, we observe a rapid decrease in the loss as the model learns to approximate the solution to the Burgers' equation. As training progresses, the rate of decrease in the loss may slow down, indicating the convergence of the optimization process. Notably, fluctuations in the loss curve may occur due to variations in the gradients and learning dynamics, highlighting the stochastic nature of the optimization algorithm.

The Fig. 06 employs the Adam optimizer, a popular choice for optimizing neural network models, to train a PINNs for solving Burgers' equation. The Adam optimizer, known for its adaptive learning rate and momentum, offers advantages in optimizing complex models by efficiently adjusting the learning rates for each parameter during training. The training process involves iteratively updating the model parameters using gradients computed from the loss function, with adjustments guided by the adaptive learning rates provided by Adam. Additionally, a piecewise decay function is utilized to adjust the learning rate throughout the training process. Specifically, for the first 1000 epochs, the learning rate is set to 0.01, followed by a decrease to 0.001 until epoch 3000, and subsequently to 0.0005 for the remaining epochs. This adaptive learning rate schedule enhances the optimization process, facilitating improved convergence and performance of the PINNs model.

After experimenting with the Adam optimizer, we're now trying out Adamax and RMSprop in Fig. 07 and Fig. 08 respectively. These are different methods for adjusting the learning rate during training, especially useful for complex tasks like solving Burgers' equation with neural networks.

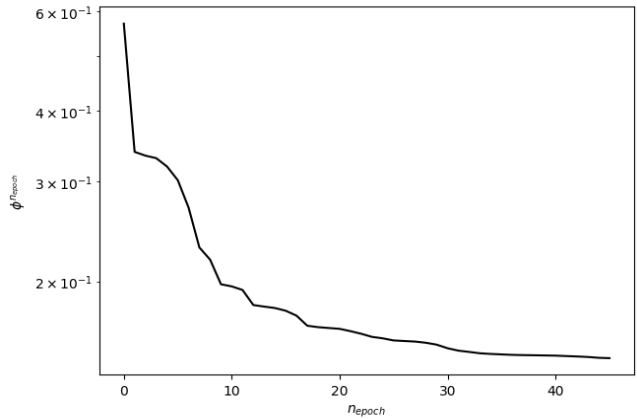

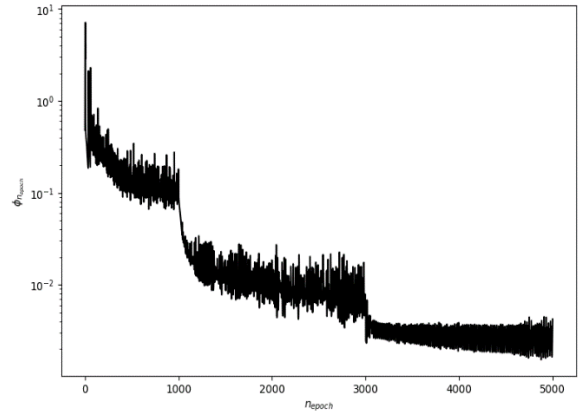

Fig.05: Loss over training Ietrations with Optimizer Adamax

Fig.06: Loss over training Ietrations with Optimizer RMS prop

Adamax is similar to Adam but is particularly effective when dealing with large amounts of data and sparse gradients. It is renowned for its adaptability and robustness in handling large-scale optimization problems. On the other hand, RMSprop adjusts the learning rate individually for each parameter, making it robust and suitable for scenarios where the problem dynamics are constantly changing.

In Fig.08, the final segment of our experiment, we examine the performance of RMSprop when training PINNs models for solving Burgers' equation. We aim to assess whether RMSprop exhibits comparable effectiveness to the other optimization methods we have explored. The 3D Visualization of Solution using the Adam Optimizer offers a graphical depiction of the optimized solution obtained from training a model using the Adam optimization algorithm. This visualization provides valuable hints into the structural characteristics and dynamic behavior of the solution in a three-dimensional space. It aids in understanding how the solution evolves over time and across different spatial dimensions, thereby enhancing our comprehension of the underlying physical phenomena described by the Burgers' equation.

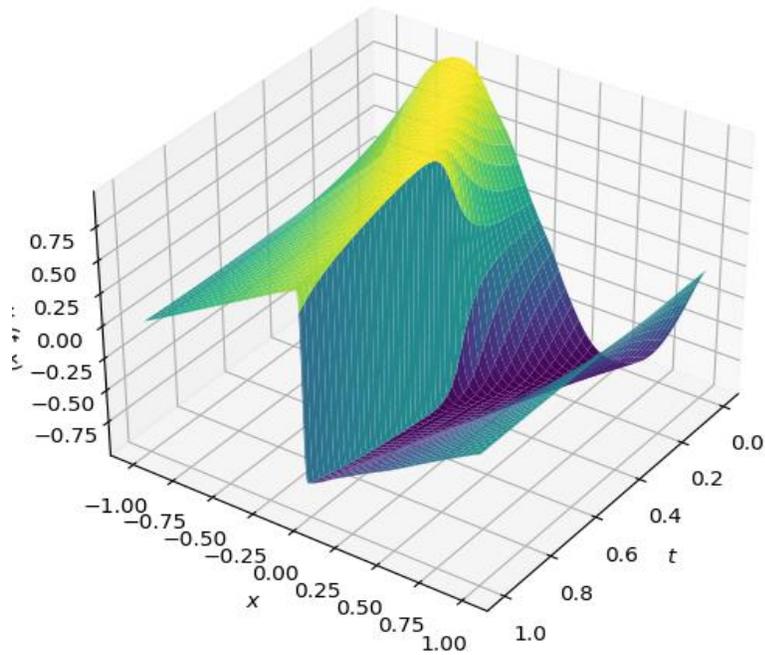

*Fig.09: Solution of Burgers' Equation in 3D*

Next, we present a series of 2D figures illustrating the solution of the Burgers equation at various time points: $t = 0.25, t = 0.5, t = 0.75, t = 0.9$. These figures serve as snapshots of the fluid velocity profile at different stages in the simulation, providing crucial insights into the evolving dynamics of the system. By visualizing the solution at multiple time instances, we gain a comprehensive understanding of how the fluid velocity field evolves over time and space.

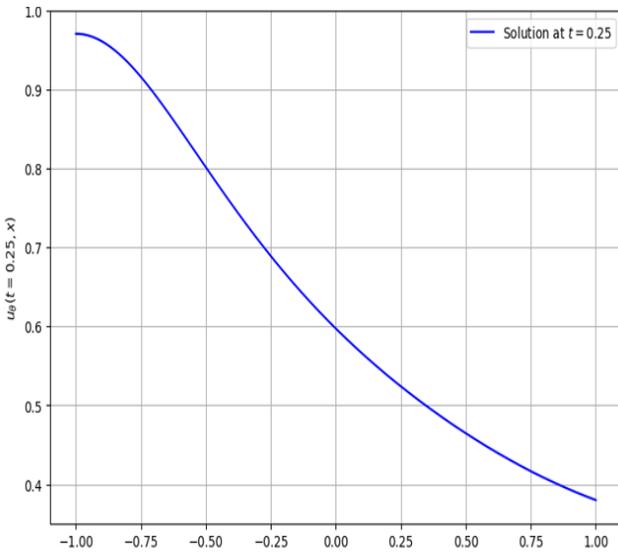

Fig.10: Solution of Burgers Equation at the time t = 0.25

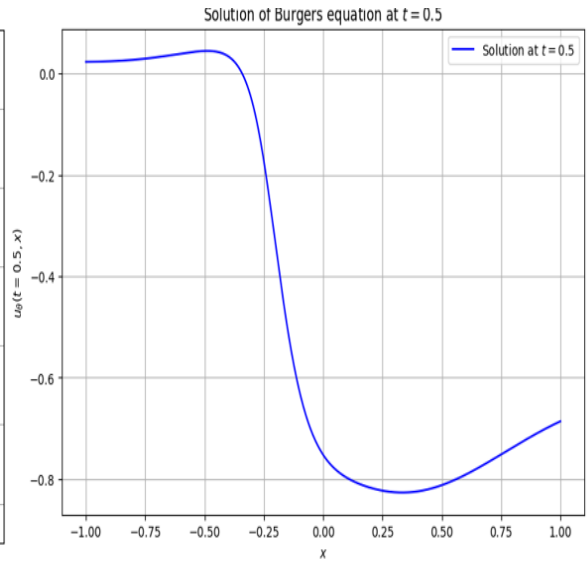

Fig.11: Solution of Burgers Equation at the time t = 0.5

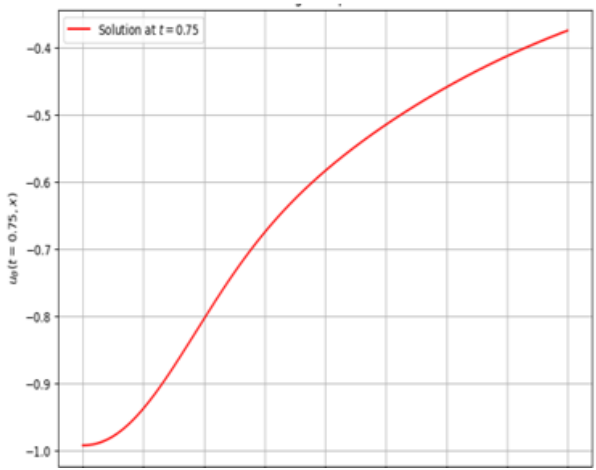

Fig.12: Solution of Burgers Equation at the time t = 0.75

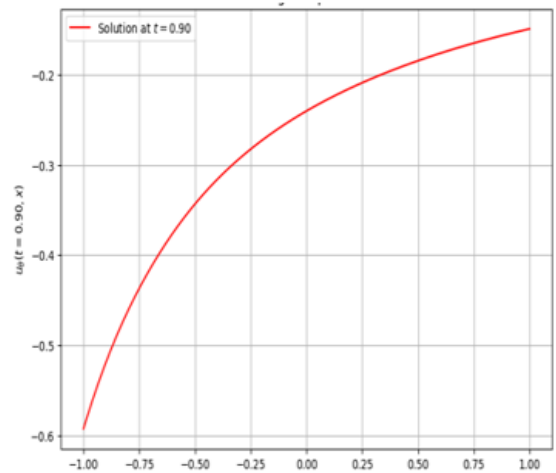

Fig.13: Solution of Burgers Equation at the time t = 1.0

The Fig.10, Fig.11, Fig12, and Fig.13 showcases the velocity distribution across spatial positions $x$ at the specified time $t$. The x-axis denotes the spatial domain, while the y-axis represents the corresponding velocity values $u_\theta$. The solution illustrates how features of the fluid flow, such as shocks or discontinuities, propagate through space over time. This propagation can reveal insights into the behavior of the fluid, such as the formation of shock waves or the evolution of turbulence. Time is a crucial factor in experiments like this, where computational resources are often limited, and researchers aim to achieve results within a reasonable timeframe. The computational time required for training neural network models using different optimization algorithms directly impacts the experimental process's efficiency and resource utilization. By analyzing and comparing the computation times of various optimization algorithms, researchers can make informed decisions regarding algorithm selection, resource allocation, and experiment planning.

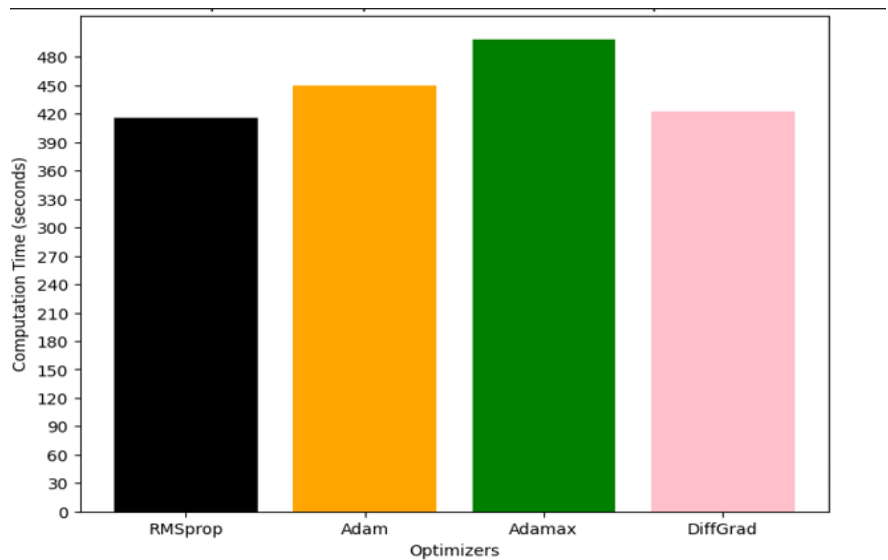

*Fig.14: Comparison of Computation time over Different Optimizers*

RMSprop and DiffGrad demonstrate relatively lower computation times, with values close to 415.96 seconds and 422.67 seconds, respectively. The graph in Fig.14 provides valuable insights into the efficiency of optimization algorithms in terms of computational performance, crucial for decision-making in model training and deployment scenarios.

**CONCLUSION**

In conclusion, our study highlights the application of Physics-Informed Neural Networks (PINNs) for solving Burgers' equation, emphasizing the importance of ensuring that the neural network learns the system's behavior while adhering to fundamental physical laws. Through a comprehensive comparative analysis, we evaluate the efficacy of various optimizers, including traditional methods like Adam, Adamax, and RMSprop, as well as the custom-defined DiffGrad optimizer, in minimizing the loss function associated with PINNs for solving Burgers' equation.

Our findings highlight the critical role of optimizer selection in optimizing the loss landscape of PINNs. Notably, while Adam exhibits lower loss compared to DiffGrad and all other optimizers, DiffGrad is found to be less computationally expensive than Adam. Despite Adam's lower loss, the computational efficiency of DiffGrad makes it an attractive option for training PINNs. Additionally, graphical representations of the solution over spatial coordinates at different time intervals provide valuable insights into the model's performance.

Overall, our results offer valuable guidance for practitioners seeking to enhance the training performance of PINNs, emphasizing the importance of considering both computational efficiency and user-friendliness when selecting optimizers. This study underscores the significance of optimizer choice in achieving optimal model convergence and efficiency in solving complex physical systems.